\documentclass{article}

\usepackage{multirow}
\usepackage{arxiv}
\usepackage{amsmath,amssymb}
\usepackage[utf8]{inputenc} 
\usepackage[T1]{fontenc}    
\usepackage{hyperref}       
\usepackage{url}            
\usepackage{booktabs}       
\usepackage{amsfonts}       
\usepackage{nicefrac}       
\usepackage{microtype}      
\usepackage{lipsum}
\usepackage{graphicx}
\graphicspath{ {./images/} }

\title{From Perceptions to Evidence: Detecting AI-Generated Content in Turkish News Media with a Fine-Tuned BERT Classifier}

\author{
Ozancan Ozdemir \\
  Bernoulli Institute\\
  University of Groningen\\
   Nijenborgh 9, 9747 AG Groningen, The Netherlands \\
  \texttt{o.ozdemir@rug.nl} \\
}

\begin{document}
\maketitle
\begin{abstract}
The rapid integration of large language models into newsroom workflows has raised urgent questions about the prevalence of AI-generated content in online media. While large-scale computational studies have begun to quantify this phenomenon in English-language outlets, no empirical investigation exists for Turkish news media, where existing research remains limited to qualitative interviews with journalists or fake news detection. This study addresses that gap by presenting the first transformer-based framework for detecting AI-generated content in Turkish news articles. We construct a labeled dataset of 3,600 articles collected from three major Turkish outlets representing distinct editorial orientations, a central media outlet, a pro-goverment media outlet and opposition aligned goverment alignment, and fine-tune a Turkish-specific BERT model (\texttt{dbmdz/bert-base-turkish-cased}) for binary classification. The model achieves 97.08\% F1 score on the held-out test set with symmetric precision and recall across both classes. We further apply the trained classifier to over 3,500 previously unseen articles from the same three sources spanning 2023–2026, providing the first computational characterization of AI-generated content patterns across Turkish media over a multi-year period. The external analysis reveals consistent classification distributions across sources and years, with mean prediction confidence exceeding 96\% and estimates that approximately 2.5\% of the news are rewritten or revised by LLM on average across the examined time frame.  To the best of our knowledge, this is the first study to move beyond self-reported journalist perceptions toward empirical, data-driven measurement of AI usage in Turkish news media. The dataset, methodology, and baseline results are intended as a foundational resource for future research at the intersection of Turkish NLP, computational journalism, and media studies.
\end{abstract}


\section{Introduction}

The public release of ChatGPT in November 2022 marked a watershed moment in the relationship
between artificial intelligence and text production since these LLMs predict the next word successfully and accurately. Within months, generative AI tools
powered by large language models (LLMs) transitioned from research curiosities to mainstream
instruments adopted across industries at unprecedented speed.  By 2025, 61\% of the global
population surveyed had used a generative AI system at least once, with weekly usage nearly
doubling from 18\% to 34\% in a single year \cite{noauthor_generative_2025}. This rapid diffusion has affected various domains including journalism. 

Due to the high-level ability of generating high-quality text, LLMs can ensure contents at a pace that far outstrips traditional methods with concerns about reduced transparency and trustworthiness. This is a remarkable opportunity for online media outlets since one of their revenue source is based on the number of visiting to their website, that can be positively correlated with the number of contents.  

One of the surveys involves AI usage in journalism in US indicate that over 80\% of newsrooms in North America now leverage some form of AI in their operations \cite{generative_ai_in_journalism}. A 2025 study of UK journalists found that 56\% use AI tools professionally on at least a weekly basis, with applications spanning story research (22\%), headline generation (16\%), idea brainstorming (16\%), and even first-draft composition (10\%) \cite{thurman_2025_adoption}. 

This transformation has also altered the journalism habits in Turkey as rest of the world as we briefly summarized in the previous paragraph. 
Journalists in Turkey appear to employ AI primarily to increase efficiency in routine newsroom tasks, including the transcription of audio recordings, the translation of foreign-language sources, the summarization of lengthy legal or technical documents, and the automated checking of grammar and style.
Beyond these operational uses, AI tools are also utilized for the production of news visuals and infographics, particularly as a means of avoiding copyright-related risks and reducing production costs, as well as for conducting large-scale data analysis to support background research. Moreover, AI-assisted rewriting practices are adopted to restructure news texts in order to address platform-specific constraints (e.g., Google News) and to optimize content for search engine visibility (SEO)\cite{coban_use_2025, demir_use_2025, yapay_zeka_gazetecilik}. In addition to this, a study by \cite{mutlu_field_2025} demonstrate that the almost 50\% journalists use AI for data analysis and visualization, and 40\% of them get use of LLM for title writing and translation. Besides, the quarter of the journalists participating in this particular study use AI for social media content and news writing. 

Despite the advantages of LLMs in news production or editing, this rapid adoption also brings several concerns raising over reduced transparency and trustworthiness. The news edited or generated by LLMs are also under the treat of bias that LLMs has inherently due to the training data. A study published in Scientific Reports examining AI-generated news content from seven major LLMs, including ChatGPT and LLaMA, found that every model produced content with substantial gender and racial biases, with systematic discrimination against women
and Black individuals \cite{Fang_Che_Mao_Zhang_Zhao_Zhao_2024}. This finding is particularly alarming given that the models were prompted with headlines from The New York Times and Reuters, outlets specifically selected for their commitment to unbiased reporting.

Another concern is caused by the hallucinations. Turkish media has a concrete example for this issue. At the end of January 2026, a newspaper in Turkey published a line-up before the Fenerbahçe's game against Aston Villa. However, the line-up had some players who used to play at Fenerbahçe before but not available in the current team squad such as Livakovic, Osayi-Samuel etc. When we ask ChatGPT to generate a line-up for Fenerbahçe, we discovered that it generates almost same players who used to play for Fenerbahçe but not being in the current team squad. The screenshots from both newspaper and ChatGPT output are given below. 

\begin{figure}[h]
    \centering
    \includegraphics[width=0.5\linewidth]{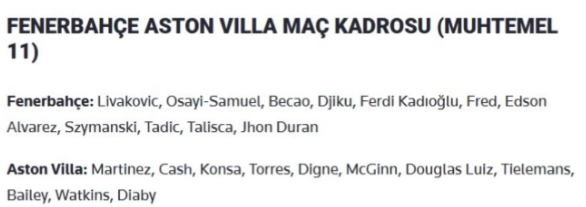}
    \includegraphics[width=0.5\linewidth]{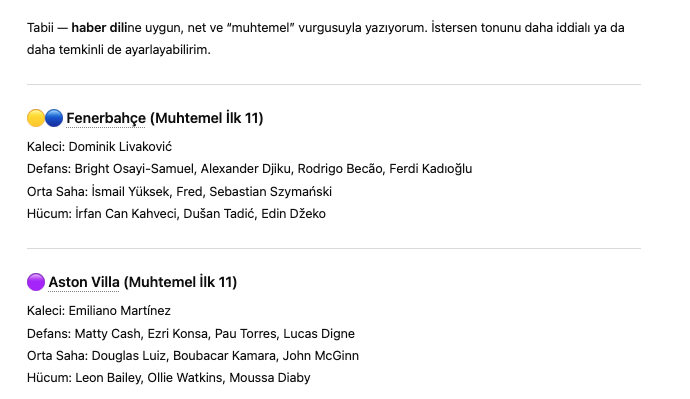}
    \caption{The line-ups in the newspaper and ChatGPT}
    \label{fig:placeholder}
\end{figure}

The bias and hallucinations are compounded by the erosion of public trust in AI-mediated news. According
to the Reuters Institute’s 2025 survey, only 12\% of respondents are comfortable with news produced entirely by AI, and only 33\% believe that journalists regularly verify AI outputs
before publication. The “comfort gap” between human-produced and
AI-produced journalism is substantial and consistent across demographics and countries in this report. 62\% of participants are comfortable with entirely human-made news, but this drops to 21\% when AI is involved with human oversight, and to 12\% when AI operates autonomously \cite{noauthor_generative_2025}. These
figures signal a fundamental tension. News organizations are rapidly adopting AI to maintain competitive output, while audiences increasingly distrust the very content these tools produce. Thus, the adoption of AI in journalism is a concept that should be taken into consideration carefully and identifying the use of AI in current news reports is of great importance.

Despite this rapidly evolving media environment, a critical gap persists in the Turkish literature. The most of the available studies, where we summarized some above, investigating the AI adoption and usage in Turkish media outlets are heavily qualitative, relying on in-depth interviews with a limited number of journalists to capture self-reported perceptions rather than empirically measuring AI presence in news content. To date, no study has applied computational methods to detect and quantify the actual rate of AI-generated content in Turkish news media. 

This gap is particularly significant for three reasons. First, Turkish is an agglutinative language
with rich morphology that poses distinct challenges for NLP systems; AI-generated or revised detection models developed for analytic languages like English cannot be straightforwardly transferred.
Second, Turkey’s unique position at the intersection of European and Middle Eastern media
traditions means that patterns of media bias may differ qualitatively from those documented
in Western contexts. Third, the practical demand for automated bias detection tools is acute,
given both the scale of AI-generated content entering the Turkish media ecosystem and the
documented erosion of editorial independence across the sector.

To the best of our knowledge, this study represents the first systematic attempt to apply BERT-based classification to AI rewritten/revised Turkish news  detection. In doing so, we aim to establish a foundational framework, encompassing a labeled dataset, a fine-tuned model, and baseline results, upon which future research can build. Specifically, this study makes the following contributions.
\begin{itemize}
    \item We construct a dataset of 3,600 news articles
sourced from three major Turkish media outlets annotated
for binary bias classification. This dataset fills a critical resource gap and provides a benchmark for future Turkish media bias research.
\item  We fine-tune a Turkish-specific
BERT model (\texttt{dbmdz/bert-base-turkish-cased}) for the bias detection task, achieving an F1 score of 97.08\% on the held-out test set and establishing the first deep learning baseline for this problem.
\item We apply the trained model to over
3,500 previously unseen news articles from the same three outlets spanning 2023–2026,
providing the first computational characterization of bias patterns across major Turkish
media sources over a multi-year period.
\item We document the complete
experimental pipeline including data collection, preprocessing, model fine-tuning, and
evaluation and thereby providing a replicable framework for researchers seeking to extend bias detection
to other under-resourced languages and media environments.
\end{itemize}
The rest of paper is organized as follows. Section 2 reviews the related literature on transformer-based text classification in Turkish media and  Section 3, Section 4 and Section 5 describe the dataset construction, and the methodology, including  model architecture, and training procedure. Section 6 presents the experimental results and discusses the findings. Section 7 covers the limitation of the current study and outlines the direction for future research. Lastly, section 8 wraps up the entire study in a few sentences.

\section{Related Works}

The detection of AI-generated or AI-rewritten news content has rapidly emerged as a critical research area following the public release of ChatGPT. The dominant approach in the literature relies on fine-tuning pre-trained transformer models for binary classification between human-written and AI-generated text. \cite{wang2023implementingbertfinetunedroberta} fine-tuned BERT and RoBERTa variants to detect ChatGPT-generated news articles, achieving 98\% precision with their optimized RoBERTa model, while \cite{wang2024aigeneratedtextdetectionclassification} reported 97.71\% test accuracy with a BERT-based detector through careful preprocessing and a 60/40 train-test split. These results confirm that transformer architectures can reliably distinguish between human and machine-generated news text, even without domain-specific architectural modifications.
Beyond standard fine-tuning, several studies have explored hybrid architectures to push detection performance further. \cite{hazim_oguz} combined RoBERTa embeddings with feedforward neural networks (RoBERTa-FNN), reaching 99.95\% accuracy on balanced datasets, and \cite{Cheng_2025} introduced role recognition and involvement measurement through their Hybrid News Detection Corpus (HNDC) and DetectEval evaluation suite, demonstrating that pre-trained language models can identify varying degrees of LLM involvement in content creation, moving beyond binary classification toward a more nuanced understanding of human-AI collaboration in news writing.
While the studies above establish the technical feasibility of AI content detection, a more recent strand of research has shifted focus from controlled experiments to real-world prevalence measurement. \cite{russell2025aiuseamericannewspapers} applied a state-of-the-art AI detector to newly published articles in American newspapers and discovered that approximately 9\% are either partially or fully AI-generated. This finding is particularly significant as it demonstrates that AI-generated content is no longer a hypothetical concern but a measurable phenomenon in mainstream journalism. However, such prevalence studies remain exclusively limited to English-language media.

In the the literacy of Turkish news related studies, the research focus has predominantly centered on fake news detection rather than AI-generated content identification. A substantial body of work has demonstrated the effectiveness of both traditional machine learning and deep learning approaches for this task. \cite{bozuyla} developed a Turkish BERT model reaching 98.5–99.9\% accuracy in COVID-19 misinformation and general fake news detection with minimal language-specific preprocessing, and \cite{koru_detection_2024} extended this by fine-tuning BERT/BERTurk with CNN and BiLSTM layers, achieving 90–94\% accuracy including evaluation on LLM-generated fake news. On the traditional machine learning side, \cite{baran} evaluated XGBoost, Random Forest, Naïve Bayes, Logistic Regression, and SVM for bilingual fake news detection, with Random Forest achieving 93.9\% accuracy for Turkish. More recently, \cite{fidan} fine-tuned LLaMA-2-7B-Turkish and Phi-3, reaching F1 scores between 0.74 and 0.99 depending on the scenario and outperforming prior deep learning models on Turkish fake news.
The effectiveness of different feature representations remains an active debate in Turkish fake news detection. \cite{Akpinar_Akpinar_Pavlovskaya_2025} demonstrated that fastText embeddings with LSTM and attention mechanisms can capture nuanced Turkish linguistic patterns, achieving 92\% accuracy and outperforming a fine-tuned Turkish BERT baseline. In contrast, \cite{altun} found that fine-tuned BERT-based classifiers outperform both TF-IDF and fastText embedding-based approaches, reaching 99.27\% accuracy. These conflicting findings suggest that optimal feature representation for Turkish text classification may be dependent on task, dataset or  a model selection.
The closest work to ours in the Turkish context is \cite{kayabas}, who trained an LSTM model to distinguish LLM-generated sentences from human-written ones in Turkish across various academic domains, achieving 0.97 accuracy and F1 score. However, this study operates at the sentence level within academic texts rather than at the document level within news articles, and it relies on a controlled experimental setup rather than application to real-world unseen data.
As the review above illustrates, the Turkish NLP literature has produced a mature and technically accomplished body of work on fake news detection. Yet a significant gap remains. No study has applied computational methods to detect and quantify AI-generated or AI-rewritten content in Turkish news media. The English-language literature has progressed from controlled binary classification experiments to real-world prevalence measurement \cite{russell2025aiuseamericannewspapers}, but this progression has not been replicated for Turkish. The present study addresses this gap by fine-tuning a Turkish BERT model for AI-generated content detection and, critically, deploying the trained classifier on over 3,500 unseen articles from three major Turkish outlets spanning 2023–2026.Thereby, it aims to enrich the studies moving from controlled experimentation to empirical measurement in the Turkish media context.

\section{Data}

We scraped 2000 randomly selected news articles published in 2021 from the three most frequently visited online newspapers in Turkey. The collected news was about current affairs happening in the selected time period. The visiting information of newspapers is obtained from the Reuters Report 2026. Since the newspapers in Turkey are politically polarized \cite{bozdag_skeptical_2022}, we selected these papers based on their political standing. To provide a relatively comprehensive insight, we opted for one newspaper from the pro-opposition, one newspaper from the pro-government, and one newspaper located in the center. Due to the privacy issues, the newspapers' names are not shared. 

The reason behind the choice of 2021 is that it is a year where we believe that all news were written by reporters and there was no help from any language models. With this choice, we aim to introduce a human-written news pattern to the model. So, these news articles constitute the human-written labels in our dataset.

ChatGPT is the most-frequently used LLM in Turkey \cite{Kemp_2025}. Thus, we get use from it to create the AI-written \& AI-corrected labeled data. We write the following prompt to ChatGPT 5.2 to rewrite or revise the gathered news. 

\texttt{You are an experienced newspaper editor of the newspaper [newspapers' name]. Your task is to rewrite or revise the provided news in line with the language, tone, and editorial style of [newspaper's name]. Follow the rules below.}
{\ttfamily
\begin{itemize}
  \item Do not change the meaning under any circumstances. Facts, dates and times, numerical values, and names of individuals and institutions must be preserved exactly.
  \item Do not alter the text length. The rewritten article should remain very close to the original length.
  \item Analyze and apply the newspaper style.
    Sentence structure, word choice, level of formality, emphasis, and news rhythm should reflect [newspaper's name] editorial language.
    \item Improve fluency, reduce repetition, simplify phrasing, and correct grammatical and punctuation errors. The output should be an editorial rewrite, not a copy.
    \item  Do not add new information, interpretation, or inference. Use only the information explicitly stated in the original text.
    \item  Do not deviate from journalistic language; avoid subjective judgments or commentary.
    \item If quotations are present, preserve all quoted material verbatim. You may revise only the sentences introducing or concluding the quotation to match the newspaper style.
\end{itemize}
}

The news generated by ChatGPT was labeled as AI-generated (revised) and added to the dataset. The following figure shows that the text length between human-written and AI-generated (revised) are close to each other, which is one indication of the verifiability of our prompt. 

\begin{figure}[h]
    \centering
    \includegraphics[width=0.5\linewidth]{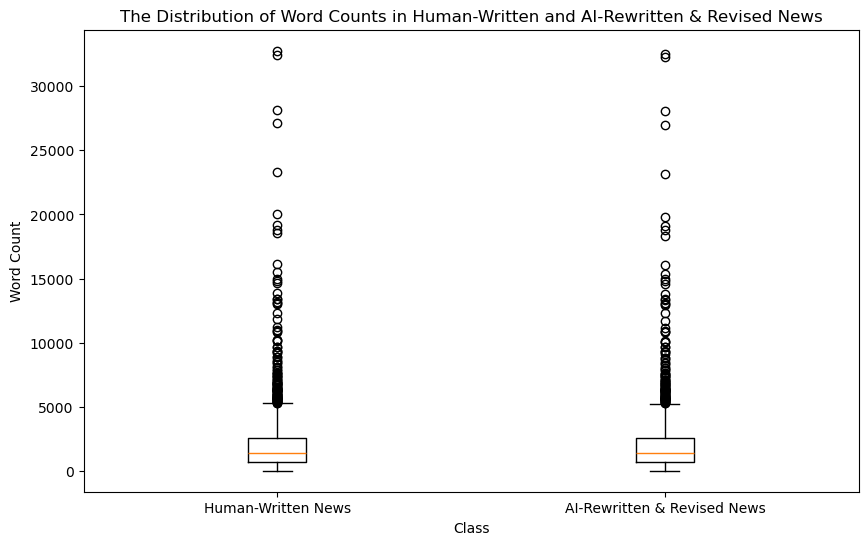}
    \caption{The Distribution of Word Counts in Human-Written and AI-Rewritten \& Revised News}
    \label{fig:placeholder}
\end{figure}
\newpage
\begin{figure}[h]
    \centering
    \includegraphics[width=0.5\linewidth]{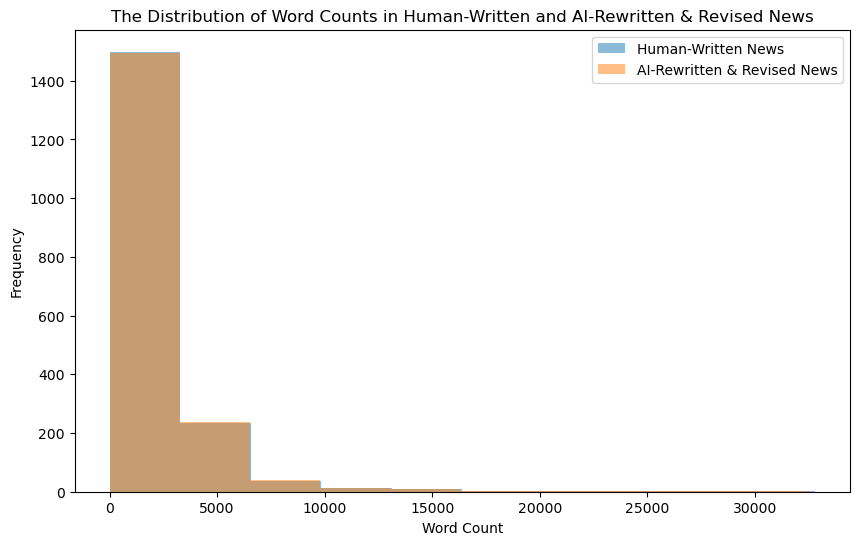}
    \caption{The Distribution of Word Counts in Human-Written and AI-Rewritten \& Revised News}
    \label{fig:placeholder}
\end{figure}

In addition to text length, similarity between texts was also measured. The primary goal was to generate news articles that are hard to distinguish from genuine news, thereby ensuring a robust and well-performing model. We used the cosine similarity measure, and it equals 0.9 on average. This value indicates that the measured set of text vectors has a high degree of semantic similarity. It means they are very closely related, and usually represent the same concept or carry the same intent. 

\section{Method}

\subsection{Bidirectional Encoder Representation from Transformers (BERT)}

Bidirectional Encoder Representation from Transformers (BERT) is a pretrained language model proposed by \cite{devlin_bert_2019} and developed based on the transformer architecture \cite{vaswani_attention_2023}. In contrast to the traditional unidirectional language models, the BERT architecture employs a bidirectional attention mechanism, and it encodes each token to attend to all other tokens simultaneously. Thereby, it achieves richer semantic representation. 

The principle of BERT mainly consists of two phases: pre-training and fine-tuning \cite{wang_ai-generated_2024}. The pre-training phase is based on two semi-supervised tasks, which are Masked Language Model (MLM) and Next Sentence Prediction (NSP).

\begin{figure}
    \centering
    \includegraphics[width=0.5\linewidth]{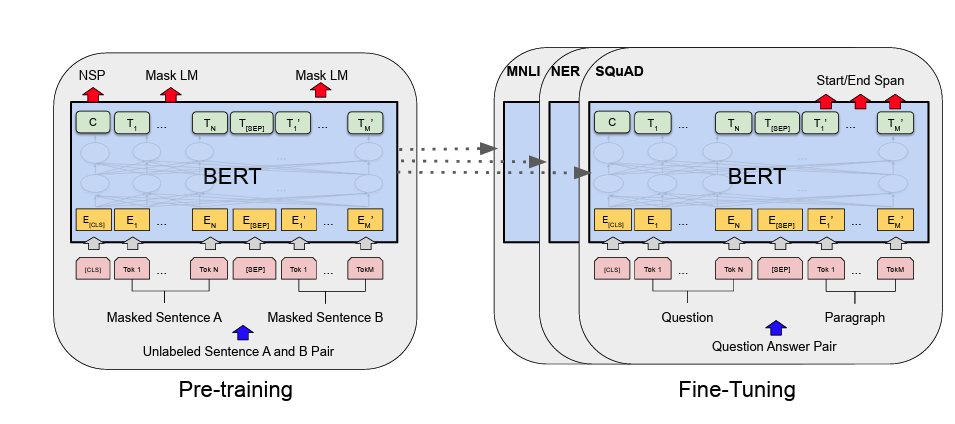}
    \caption{The Pre-training and fine-tuning procedures of BERT \cite{devlin_bert_2019}}
    \label{fig:placeholder}
\end{figure}

In MLM, a random portion of the input sequence, roughly 15\%, is masked, and the model learns to predict them from their bidirectional context. This forces the model to model the contextual information and learn better word representation\cite{wang_ai-generated_2024}. On the other hand, the model learns to decide whether two sentences are consecutive in the original corpus in NSP, enabling understanding the correlation between sentences that brings better insight for text sequence. In short, the dual-objective pre-training procedures produce contextual word embeddings that encode both synthetic and semantic information, enabling effective transfer learning across a wide range of tasks.
The BERT model can be adapted to a supervised learning task after adding one layer, combined with labeled data. 
The base model architecture consists of 12 Transformer encoder layers, 768 hidden dimensions, 12 self-attention heads, and approximately 110 million parameters. Each encoder layer comprises a multi-head self-attention sub-layer followed by a position-wise feed-forward network,
with layer normalization and residual connections applied after each sub-layer.

\subsection{Turkish BERT}

While the original BERT models were pre-trained primarily on English corpora, language-specific models have been shown to significantly outperform multilingual variants on monolingual
tasks \cite{nozza_what_2020, virtanen_multilingual_2019}. For Turkish NLP tasks, we employ
\texttt{dbmdz/bert-base-turkish-cased}, a Turkish-specific BERT model developed by the
Munich Digitization Center (MDZ) at the Bavarian State Library in collaboration with the Technical
University of Munich \cite{noauthor_stefan-itturkish-bert_nodate}.
This model was pre-trained on a large-scale Turkish corpus comprising approximately 35 GB
of text, including data from the Turkish subset of OSCAR \cite{suarez_asynchronous_2017}, the Turkish
Wikipedia, and the OPUS corpus \cite{noauthor_tiedemann_nodate}. Crucially, the model utilizes a Turkish-specific
vocabulary of 32.000 WordPiece tokens generated through a dedicated tokenization procedure, which preserves the morphological richness and agglutinative structure of Turkish.
The cased variant preserves capitalization information, which is particularly important for
Turkish due to the distinction between dotted and undotted letters (I/ı and İ/i). This model has
demonstrated state-of-the-art performance across various Turkish NLU benchmarks, including
sentiment analysis, named entity recognition, and text classification tasks.

\subsection{Fine-Tuning for Text Classification}

Transfer learning via fine-tuning has become the de facto paradigm for adapting pre-trained
language models to task-specific objectives \cite{howard_universal_2018}.In the fine-tuning framework, the pre-trained model weights serve as an initialization, and the entire
model, including both the pre-trained encoder and a newly added task-specific classification head, is updated end-to-end on the labeled downstream dataset.
For the binary text classification task in this study, we append a single linear classification
layer on top of the [CLS] token representation from the final encoder layer:

\begin{equation}
    \hat{y}=\operatorname{softmax}\left(\mathbf{W} \cdot \mathbf{h}_{\mathrm{[CLS]}}+\mathbf{b}\right)
\end{equation}

where $\mathbf{h}_{\text {[CLS]}} \in \mathbb{R}^{768}$ is the pooled output representation of the [CLS] token, $\mathbf{W} \in \mathbb{R}^{2 \times 768}$ is the weight matrix of the classification head, and $\mathbf{b} \in \mathbb{R}^2$ is the bias vector. The softmax function
maps the logits to a probability distribution over the two classes.
The classification loss is computed using the cross-entropy objective with label smoothing \cite{noauthor_rethinking_nodate}.

\begin{equation}
    \mathcal{L}=-\sum_{i=1}^N\left[\bar{y}_i \log \left(\hat{y}_i\right)+\left(1-\tilde{y}_i\right) \log \left(1-\hat{y}_i\right)\right]
\end{equation}

where $\bar y = (1-\alpha).y_i + \alpha/K$ is the smoothed label, $\alpha = 0.1$ is the smoothing factor, $K=2$ is the number of classes, and n is the number of training samples. Label smoothing serves as a regularization mechanism that prevents the model from becoming overconfident in its predictions, thereby mitigating overfitting.

\section{Data Pre-Processing and Model Training}

\subsection{Data Cleaning}

Before modeling, the collected data are subjected to cleaning operations. Since news items are randomly selected, some overlap, resulting in duplicates. To get rid of these, we filtered the news based on the date and removed the duplicates by keeping an equal distribution between the journals. In the end, we have 1800 news items left, almost equally distributed across the three newspapers.  In data exploration, we observed that some scraped news items also contain headlines that distort the narrative. We also removed them based on the text pattern. We lowercased every news item and removed the unnecessary punctuation and stopwords. Last, we cleaned the news agency name placed at the end of the news texts from one newspaper particulary since it may lead to a bias in inference. 

\subsection{Data Splitting}

The cleaned dataset was divided into training, validation, and test subsets using a stratified split
to preserve label proportions across all partitions. The allocation followed an 80/10/10 ratio,
yielding 2,880 training samples, 360 validation samples, and 360 test samples. Stratification
was enforced using the class labels as the stratifying variable to ensure balanced representation,
which is critical for training stability and unbiased evaluation \cite{noauthor_study_nodate}. The distribution of classes in each dataset is given below. 

\begin{table}[]
    \centering
    \begin{tabular}{|c|c|c|}
    
         \hline \textbf{Data} & \textbf{Human Written} & \textbf{AI Rewritten or Revised} \\
         \hline  Train & 0.5 & 0.5 \\
         \hline Validation &0.499 & 0.501 \\
         \hline Test & 0.501 & 0.499 \\
         \hline
    \end{tabular}
    \caption{The Class Distribution Across Datasets}
    \label{tab:placeholder}
\end{table}

\subsection{Tokenization}

All text inputs were tokenized using the WordPiece tokenizer associated with the \texttt{dbmdz/bert-base-turkish-model}. Sequences were padded or truncated to a maximum length of 512 tokens, which corresponds
to BERT’s architectural limit. This maximum length was selected to preserve as much
contextual information as possible from news articles, which tend to be substantially longer
than the more commonly used 128 or 256 token lengths.

\subsection{Optimization and Regularization}
The model was optimized using the AdamW optimizer \cite{loshchilov_decoupled_2019} with a
learning rate of $2\times 10^{-5}$ and a weight decay coefficient of 0.01. A linear learning rate scheduler 
with warm-up was employed, where the learning rate linearly increases from zero during the
first 10\% of training steps (warm-up phase) and then linearly decays to zero over the remaining
steps. Gradient clipping with a maximum norm of 1.0 was applied to prevent gradient explosion.
To mitigate overfitting, several regularization strategies were employed.
\begin{itemize}
    \item dropout with a
probability of 0.2 was applied to both the hidden states and the attention weights (increased
from the default 0.1)
\item label smoothing with a factor of 0.1 was used to soften the target
distribution.
\item early stopping with a patience of 2 epochs was implemented based on the
validation F1 score.
\end{itemize}

The model achieving the highest validation F1 score across all epochs was
selected as the final model.  Mixed-precision training (FP16) was utilized to accelerate training and reduce GPU memory
consumption without sacrificing model quality \cite{micikevicius_mixed_2018}.
The following table summarizes the complete set of hyperparameters used during fine-tuning.

\begin{table}[h]
    \centering
    \begin{tabular}{|c|c|}
 \hline \textbf{Hyperparameter} & \textbf{Value} \\
\hline Pre-trained model & \texttt{dbmdz/bert-base-turkish-cased} \\
\hline Maximum sequence length & 512 tokens \\
\hline Number of labels & 2 (binary classification) \\
\hline Optimizer & AdamW \\
\hline Learning rate & $2 \times 10^{-5}$ \\
\hline Weight decay & 0.01 \\
\hline LR scheduler & Linear with warm-up \\
\hline Warm-up ratio & 0.1 (10\% of total steps) \\
\hline Maximum gradient norm & 1.0 \\
\hline Maximum epochs & 6 \\
\hline Batch size (per device) & 8 \\
\hline Gradient accumulation steps & 2 \\
\hline Effective batch size & 16 \\
\hline Hidden dropout probability & 0.2 \\
\hline Attention dropout probability & 0.2 \\
\hline Label smoothing factor & 0.1 \\
\hline Early stopping patience & 2 epochs \\
\hline Early stopping metric & Validation F1 score \\
\hline Precision & FP16 (mixed precision) \\
\hline Random seed & 42 \\
\hline
    \end{tabular}
    \caption{Hyperparameter configuration for BERT fine-tuning}
    \label{tab:placeholder}
\end{table}

\subsection{Evaluation}

Since it is a classification task, a model's performance was evaluated across four conventional classification measures. Accuracy, precision, recall, and F1 score. Given the balanced dataset, accuracy provides a reliable insight into the model's performance, but the F1 score was preferred as the primary metric for the model selection phase since it provides a more comprehensive assessment of classifier performance, particularly in scenearios where both false positives and false negatives carry significant costs. \cite{noauthor_systematic_nodate} Besides these four measures, we also evaluated the reliability score for predictions so that we could measure how confident the model is in prediction.

\begin{equation}
\text {Accuracy} =\frac{TP+TN}{TP+TN+FP+FN} 
\end{equation}

\begin{equation}
\text{Precision} = \frac{TP}{TP+FP}
\end{equation}

\begin{equation}
\text{Recall} = \frac{TP}{TP+FN}
\end{equation}

\begin{equation}
F_{1} = 2 \cdot \frac{\text{Precision}\cdot\text{Recall}}{\text{Precision}+\text{Recall}}
\end{equation}

\subsection{Implementation Details}
Implementation Details
All experiments were conducted on Google Colaboratory using a single NVIDIA Tesla T4 GPU
(16 GB VRAM). The implementation was based on the Hugging Face \texttt{transformers} library
(v5.0) \cite{wolf_huggingfaces_2020} with PyTorch (v2.7) as the deep learning backend. The \texttt{Trainer} API
was used for training orchestration, including automated logging, checkpoint management, and
early stopping. The total training time was approximately 18 minutes

\section{Results and Discussion}

The fine-tuning of the Turkish BERT was performed over six epochs, with the model exhibiting rapid convergence during the initial training phase. 

\begin{figure}[h]
    \centering
    \includegraphics[width=0.5\linewidth]{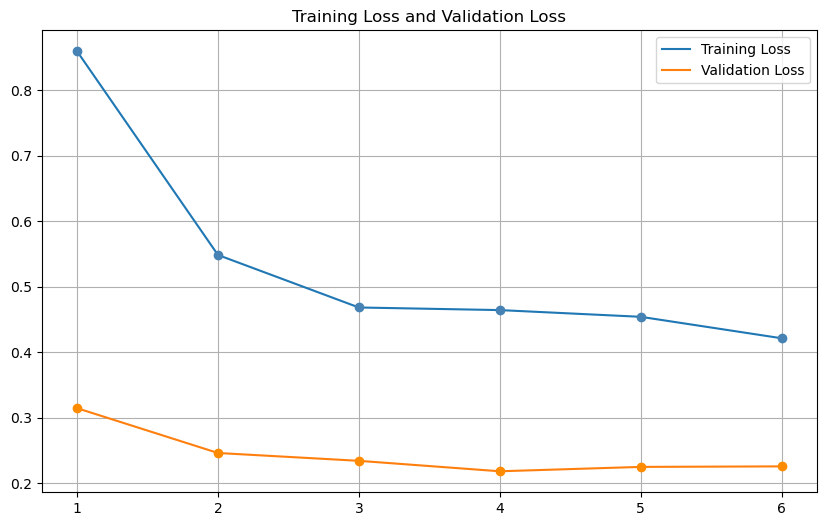}
    \caption{Training Loss and Validation Loss Over Epochs}
    \label{fig:train}
\end{figure}

The training loss decreased substantially from 0.8597 in the first epoch to 0.422 in the sixth,
reflecting effective parameter adaptation and efficient
convergence. Notably, the validation loss reached its minimum of 0.218 at epoch 4, after which a marginal increase was observed in epochs 5 and 6 (0.225 and
0.226, respectively), indicating the onset of overfitting. The best-performing model, selected
based on the highest validation F1 score of 0.991, was saved at epoch 4. 

We also test the trade-off between precision and recall in the training. In epoch 1, the model achieved high precision (0.985) but comparatively lower recall
(0.8994), suggesting a conservative classification strategy in the early stages. By epoch 3, this
asymmetry was resolved, with both metrics converging to roughly 0.99, indicating that the model
developed a more balanced decision boundary as fine-tuning progressed.

\begin{table}[]
    \centering
    \begin{tabular}{|c|c|c|c|}
         \hline \textbf{Accuracy} & \textbf{F1} & \textbf{Precision} & \textbf{Recall}  \\
         \hline 0.991098	& 0.990596	& 0.987500	& 0.993711\\
         \hline
        
    \end{tabular}
    \caption{The Best Performing Model's Metrics on Training Data}
    \label{tab:placeholder}
\end{table}

Although the training period exhibits that the model is learning successfully, an unexpected behaviour is observed, where the validation loss is below the training loss. This situation arises from the strong regularization strategy employed, specifically the combination of Dropout and Label Smoothing. During training, Dropout reduces the model's capacity, and Label Smoothing prevents the loss from reaching zero by introducing soft targets. In contrast, during inference/validation, the full model capacity is utilized without dropout, and performance is evaluated against ground-truth labels. 
This confirms that the model is effectively regularized against overfitting.

It appears from the Figure \ref{fig:train} that the employed strategy to reduce the overfitting has contributed to the stability of the training
process. The training loss consistently decreased across all six epochs, while the validation loss
stabilized after epoch 3, exhibiting only marginal fluctuations. This pattern suggests that the
combined regularization approach was effective in moderating the rate of overfitting, though it
did not entirely prevent it. The saved model was evaluated on the test set including 360 samples. The detailed evaluation is presented below. 

\begin{table}[h]
    \centering
    \begin{tabular}{|c|c|c|c|}
         \hline \textbf{Accuracy} & \textbf{F1} & \textbf{Precision} & \textbf{Recall}  \\
         \hline 0.9792	& 0.9708	&  0.9938	& 0.9878\\
         \hline   
    \end{tabular}
    \caption{Model's Performance on Test Data}
    \label{tab:placeholder1}
\end{table}

The model achieved an overall accuracy of 97.92\% on the test set, with perfectly symmetric
performance across both classes, for each F1 with 0.9708. The near-identical precision and recall
values for both classes confirm that the model does
not exhibit systematic bias toward either class. 
Compared to the validation set performance at epoch 4 (F1 = 0.9941), the test set F1 score
of 0.9708 represents a modest decrease of 2.33 percentage points, suggesting a slight degree of
overfitting to the validation distribution. Nevertheless, the test performance remains robust and
indicates strong generalization within the domain of the training corpus.

\subsection{External Validation}

Since the main aim of this study is to detect the news rewritten or revised by ChatGPT and calculate the estimated proportion, we applied the fine-tuned model to the external datasets. Thereby, we aim to assess the generalization capability of the model beyond the training distribution. To this end, we gathered 300 randomly selected news items from each of the years 2023, 2024, 2025, and 2026 from the same media outlets. The average proportion over the years across the three newspapers given below. 

\begin{figure}[h]
    \centering
    \includegraphics[width=0.5\linewidth]{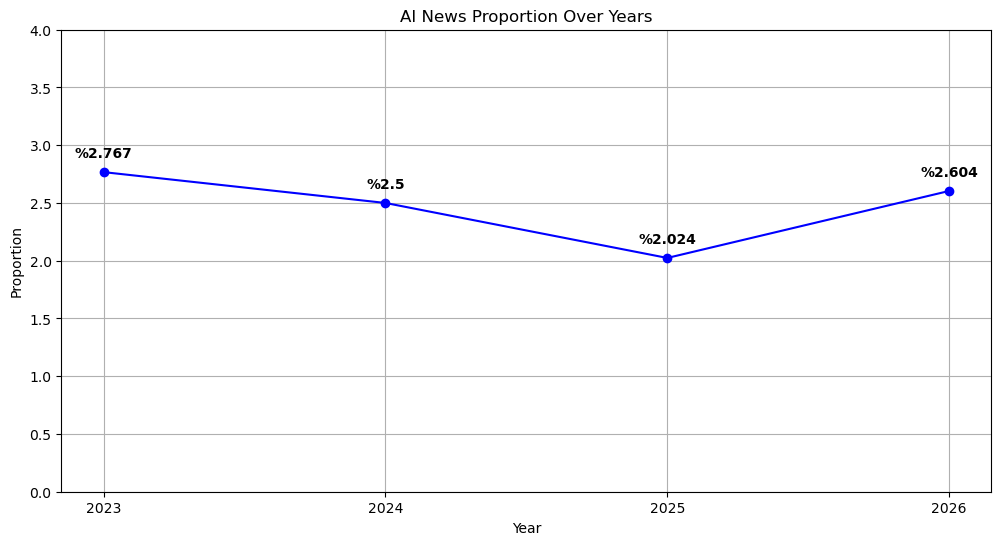}
    \caption{The proportion of the news rewritten or revised by LLM}
    \label{fig:placeholder}
\end{figure}

The table below shows the model prediction for each class with the corresponding reliability score. 

\begin{table}[htbp]
\centering
\caption{External validation results: predicted class distributions and mean confidence scores across media sources and years.}
\label{tab:external_results}
\begin{tabular}{lcccc}
\toprule
\textbf{Source} & \textbf{Year} & \textbf{Human-Written (\%)} & \textbf{AI Rewritten or Revised (\%)} & \textbf{Mean Conf. (\%)} \\
\midrule
\multirow{4}{*}{Central Newspaper}
 & 2023 & 96.7 & 3.3 & 96.5 \\
 & 2024 & 96.7 & 3.3 & 96.3 \\
 & 2025 & 98.7 & 1.3 & 96.5 \\
 & 2026  & 97.3 & 2.7 & 96.4 \\
\midrule
\multirow{4}{*}{Pro-Opposition Newspaper}
 & 2023 & 98.3 & 1.7 & 96.4 \\
 & 2024 & 98.3 & 1.7 & 96.3 \\
 & 2025 & 98.0 & 2.0 & 96.5 \\
 & 2026 & 98.0 & 2.0 & 96.5 \\
\midrule
\multirow{4}{*}{Pro-Goverment Newspaper}
 & 2023 & 96.4 & 3.6 & 96.4 \\
 & 2024 & 97.5 & 2.5 & 96.1 \\
 & 2025 & 97.2 & 2.8 & 96.2 \\
 & 2026 & 97.3 & 2.7 & 96.2 \\
\bottomrule
\end{tabular}
\end{table}

The high reliability score across the media outlets and classes indicates that the model predicts with high confidence. Alongside this, the successful training and validation performance confirms that LLM, ChatGPT, in the context of this study, has been used in the selected Turkish media outlets since the release of ChatGPT. 

We estimate that the 2.5\% of the news are rewritten or revised by ChatGPT over the last 4 year on average. In Figure \ref{fig:placeholder}, the highest usage happened in 2023, a year after the release of ChatGPT. Since then, the usage has dropped until 2026. This year, although we only covered the news in January 2026, the usage of LLM in news production has risen and reached the level of 2023.  

If we look into the table above for a more detailed investigation, the model consistently classified the overwhelming majority of external news articles as human-written, with proportions ranging from 96.4\% to 98.7\%. The model’s predictions were made with high certainty regardless of the source or temporal period, showing that a consistency of prediction patterns across the three media outlets despite their differing editorial orientations. While the differences are modest, the model captures subtle lingustic or content-based variations across sources, reinforcing the robustness of
the model’s learned representations. The absence of significant year-over-year trends indicates
that the classification patterns are not driven by transient events or topical shifts but rather by
stable underlying characteristics of the news texts. This result may also indicate an important empirical finding under the assumption that if the external data used in the study genuinely reflect the prevalence of human-written text. 

Besides, several patterns merit attention. First, pro-opposition outlet exhibited the highest human-written news proportion
across all years, while pro-goverment newspaper showed the lowest, with central media outlet falling between the two. Second, no clear temporal trend was observed within any source; the year-over-year variations remained within a narrow band of approximately 2 percentage
3 points. Third, the consistency of the mean confidence scores (96\%) across all conditions suggests
that the model’s decision boundary is stable and not subject to substantial uncertainty in
the external domain.

\section{Limitations and Future Studies}

This study has several limitations that should be acknowledged. First, only three media outlets, although the polarization in media was taken into consideration in selection, may not be fully representative of Turkish media. However, the difficulty of gathering news published in 2021 is another reason for this selection, which is a limitation of this study. If this difficulty is overcome, the trained model could provide better and more comprehensive insights. The dataset of 3600 cleaned samples, while sufficient for demonstrating feasibility, is modest by contemporary fine-tuning standards. In the production of AI rewritten or revised texts, the generated results may not be consistent with the current editorial style in the selected newspapers. Second, BERT's 512-token limit necessitates truncation of longer articles, potentially discarding informative content given that the 75th percentile of text length in the corpus exceeds 2500 characters. Third, the external validation lacks ground-truth labels, meaning that the model's out-of-domain accuracy cannot be directly quantified; the consistent human-written class dominance ($\geq$96\%) across all sources and years may reflect either the genuine distribution of bias in Turkish media or a systematic generalization limitation. Finally, the absence of baseline comparisons with traditional classifiers (e.g., SVM, Logistic Regression) and alternative transformer models (e.g., multilingual BERT) limits the contextualization of the reported performance.

Future work should prioritize three directions. The most impactful extension would be the creation of human-annotated external test sets, sampling 100,200 articles per source and having them labeled by multiple independent annotators, to establish rigorous cross-domain performance metrics. Equally important is the 
gathering more news items from different news categories such as sports, life, health, and investigating AI-generated or revised context proportion under these categories, combined with the time range. 
Another enrichment is the inclusion of baseline comparisons with both traditional machine learning models and alternative pre-trained language models, supplemented by stratified $k$-fold cross-validation for robust performance estimation. Beyond these methodological refinements, the binary classification scheme should be extended toward multi-class and multi-label formulations that capture distinct bias dimensions (framing, selection, ideological), and long-context architectures such as Longformer \cite{beltagy2020longformerlongdocumenttransformer} or sliding-window strategies should be explored to address the truncation limitation. A systematic error analysis using interpretability techniques (LIME, SHAP) would further illuminate the model's decision boundaries and failure modes, strengthening both the scientific contribution and the practical utility of the classifier for real-world media monitoring applications.

\section{Conclusion}

This study presents, to the best of our knowledge, the first application of transformer-based classification to  AI rewritten or revised news detection in
Turkish news media by fine-tuning a Turkish-specific BERT model on a purpose-built dataset of 3,600 articles from three editorially distinct sources. We
demonstrate that pre-trained language models can effectively distinguish AI rewritten/revised texts from human-written reporting in Turkish, achieving an F1 score of 97.08\% on the held-out test set with symmetric
performance across classes. The large-scale external application of the model to over 3,500
unseen articles spanning 2023–2026 reveals consistent cross-source and temporally stable prediction patterns, providing the first computational characterization of bias distributions in the
Turkish media ecosystem. Taken together, these contributions, the first labeled data for AI rewritten/revised news in Turkish media, the first deep learning baseline for this task, and the first multi-year exploratory analysis, establish a foundational framework upon which future research can build, extending to richer annotation schemes, additional genres, and multilingual comparative studies. 

Beyond its technical contributions, this study carries a broader implication. At a time when LLM-generated content is rapidly and quietly entering newsrooms worldwide, the absence of transparency mechanisms poses a direct threat to public trust in journalism. Our findings demonstrate that even modest-scale, language-specific efforts can yield meaningful first steps toward computational media accountability in under-resourced languages. We strongly recommend that any news article generated, rewritten, or revised with the assistance of a language model be published with a clear label disclosing the nature and extent of AI involvement. We believe that this practice  would protect readers' right to informed consumption, align with the growing international consensus on AI transparency in media \cite{noauthor_generative_2025}, and preserve the credibility that remains journalism's most valuable and most fragile asset.

\section{Acknowledgement}

I would like to thank  Sercan Karakas for this help in the arXiv endorsement.

\bibliographystyle{unsrt}  
\bibliography{references}  






\end{document}